\documentclass[letterpaper, 10pt, conference]{ieeeconf}

\IEEEoverridecommandlockouts
\usepackage{graphicx}
\usepackage{amsmath, amssymb}
\usepackage{booktabs}
\usepackage{multirow}
\usepackage{wrapfig}
\usepackage{cuted}        
\usepackage[table]{xcolor}   
\usepackage[hidelinks]{hyperref}

\AtBeginDocument{%
  \let\CLAPoldref\ref
  \renewcommand{\ref}[1]{\textcolor{red}{\CLAPoldref{#1}}}%
}

\title{\LARGE \bf
CLAP: Closed-Loop Alignment with Pressure for\\ Precise Suction Manipulation
}

\IEEEoverridecommandlockouts

\author{
Yixian Zou$^{1}$,
Chongyang Xu$^{2}$,
Yuling Xin$^{1}$,
Ziliang Feng$^{2}$,
Fanman Meng$^{1}$,
Shuaicheng Liu$^{1*}$%
\thanks{$^{1}$University of Electronic Science and Technology of China.}%
\thanks{$^{2}$School of Aeronautics and Astronautics, Sichuan University.}%
\thanks{Yixian Zou: zouyixian@std.uestc.edu.cn.}%
\thanks{$^{*}$Corresponding author: Shuaicheng Liu, liushuaicheng@uestc.edu.cn.}%
}

\begin{document}
\bstctlcite{IEEEexample:BSTcontrol}
\IEEEaftertitletext{\vspace{-50pt}}
\maketitle

\begin{strip}
  \centering
  \includegraphics[width=0.92\textwidth]{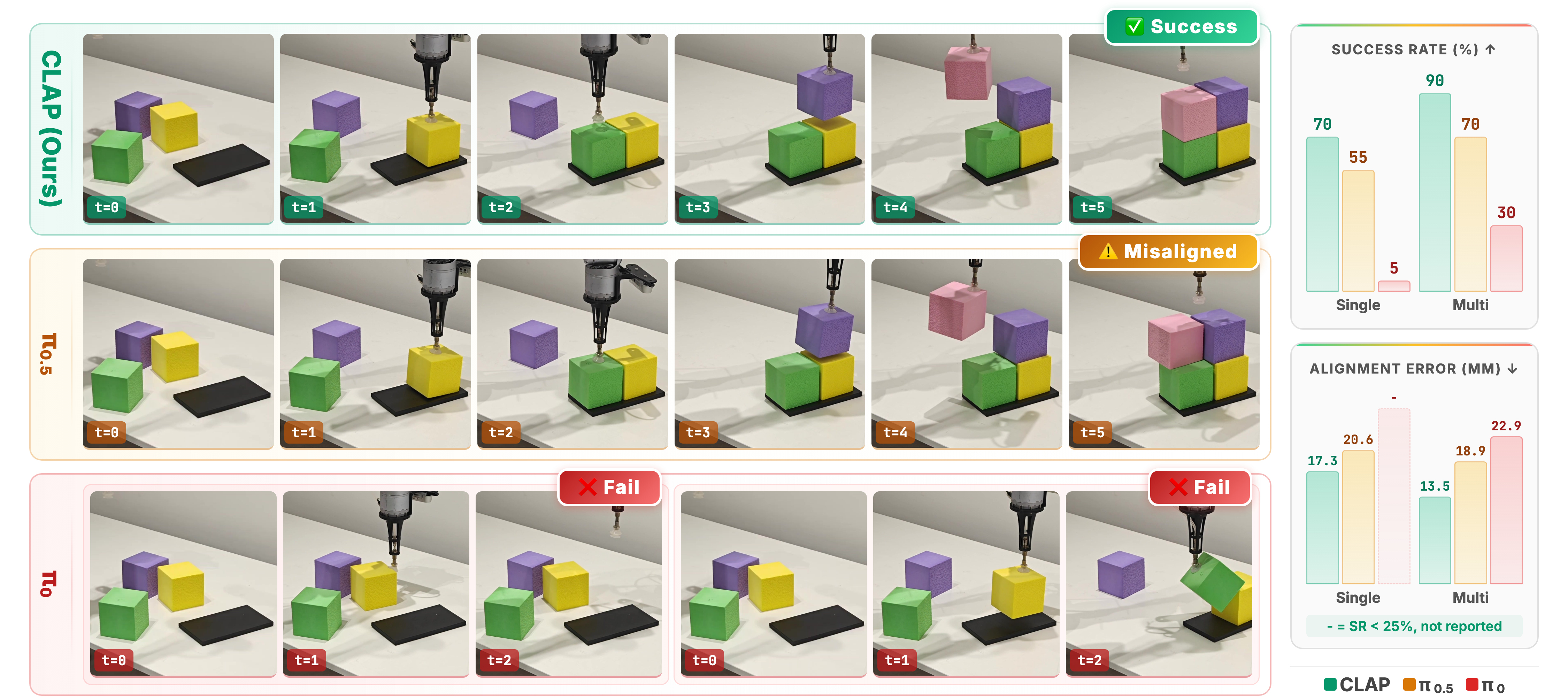}
  \par\vspace{2pt}
  \refstepcounter{figure}\label{fig:teaser}%
  \parbox[t]{\textwidth}{\footnotesize \noindent \textbf{Fig.~\thefigure.~~ CLAP against two VLA baselines.}~~ The same
  $P_4$ task in several colours: CLAP completes both stacks, $\pi_{0.5}$ places
  every block but ends misaligned, $\pi_0$ fails twice. Right: success rate and alignment error, single- and multi-task. Error is
  omitted where too few trials succeed.}
\end{strip}

\thispagestyle{empty}
\pagestyle{empty}

\begin{abstract}
Stacking and palletising demand precise placement: error left in one layer is
inherited by the next, and a flat pad offers no feature to funnel a wrong pose
into the right one. Top-down suction suits such
dense arrangements, and suction has already been brought into
vision-language-action (VLA) policies. What that work does not report, however,
is a policy conditioned on a measured vacuum signal, or one that uses it to
abandon an action already under way. Vision does not settle the question here,
because at the moment it matters the cup and the face it holds occlude
each other. We present CLAP, which makes the attachment state observable
through a pressure module tapped into the vacuum line. The decoded reading
replaces the suction command in the policy's proprioception, is fused with the
visual features, and terminates the open-loop execution window so that the
policy re-infers from a fresh observation. For data, we record goal-state
disassembly on the physical robot and reverse the joint-state sequence offline,
without a simulation replay. Targeted phase demonstrations, $8.3\%$ of the
training frames, cover the suction transitions and the configurations an
interrupted grasp leaves behind. On a real Unitree Z1, one multi-task
checkpoint reaches $96.67\%$ average success in both colour settings, $16.67$
and $10.00$ points above the strongest baseline, its monochromatic four-block
successes averaging $15.92$~mm of error. Four ablation settings fall $5.00$ to
$13.33$ points short. We will release code and trained weights.
\end{abstract}

\section{Introduction}
\label{sec:intro}

Robots that stack and palletise are held to a placement tolerance that has
little to do with reaching. Error in one layer is inherited by the next, and a
stack that leans far enough collapses. Shelving, packing and machine tending
share this tolerance~\cite{romero2025palletizing}. Unlike peg-in-hole assembly,
nothing funnels a slightly wrong pose into the right one. A block comes to rest
on a flat pad or on another block, steered only by the policy's own estimate.
Nor is this a single alignment. In the four-block task the arm places four times in sequence,
each residual entering the initial condition of the next. Classical pipelines split the task into
perception, planning and control
stages~\cite{romero2025palletizing}, letting error cross module
boundaries and needing re-engineering per geometry.
Vision-language-action (VLA) models collapse the three into one
policy~\cite{pmlr-v229-zitkovich23a,BlackK-RSS-25,intelligence2025pi05visionlanguageactionmodelopenworld},
one checkpoint covering configurations that would otherwise need separate
programs. Their loop, though, closes without any direct report of the contact
outcome: a strong
VLA reaches the target region and fails to seat the block
(Fig.~\ref{fig:teaser}).

A suction cup is essential here because it makes the placement geometrically
possible. A parallel jaw needs side clearance to open and close,
and in a dense arrangement that clearance is where the already-placed blocks
are, so setting the second block down disturbs the first. Thin, fragile or
tightly packed objects are not side-graspable at all. Our Z1's stock gripper makes the point
concrete: a single jaw rotating through $90^\circ$, and an opening we measured at
roughly $50$~mm against $80$~mm blocks. Suction need not replace other end-effectors. For this
class of placement it is the one that works. It has already been brought into
VLAs: VacuumVLA with a combined jaw-and-cup
end-effector~\cite{zhou2025vacuumvla}, GVLA with the cup as one of several
interchangeable gripper types~\cite{zhang2026gvla}.
Between them they establish what a cup can \emph{pick up}. We ask whether it can
\emph{put things down accurately}, which turns on a different signal. Neither
reports conditioning a policy on a measured vacuum reading, nor using one to
abandon an action already under way, even though the vacuum line carries exactly that reading from a component the
end-effector already has.

Three obstacles stand in the way. \emph{The data is poorest where it matters.}
Teleoperating the last few millimetres is where an operator is weakest: the hand
wavers, the operator pauses and retries, and error piles up at the end of the
trajectory, precisely the contact phase the policy must learn from.
\emph{The policy cannot tell whether its command took effect.} Our cameras do
not resolve it, and not for want of a better one: at the moment it matters, the
cup and the face it holds occlude each other. The consequence shows in the
baselines: the one field that could report the outcome carries the suction
command instead, so after a missed grasp it still reads \emph{suck}, and the
policy is told it holds a block it never picked up. And because execution proceeds in an open-loop window,
a policy that could notice afterwards may notice too late. \emph{Variation is
missing where it matters.} One high-quality trunk of demonstrations is thin on
what decides the contact phase: where on the face the cup lands, from which direction the arm
approaches, how the blocks were laid out.

CLAP answers the three in order. We let the observation report what the cup
actually did: the suction state becomes an observable, entering the policy's
proprioception in place of the command usually recorded there, taking part in
visual reasoning, and serving as the abort condition of an action block during
execution. Two demonstration strategies then make that mechanism learnable. We
present this as a chain rather than a set of pillars: pillars stand in parallel,
so removing one leaves the others holding, while links in series each bound what
the rest can deliver. Across four ablation settings the observed drop is $5.00$
to $13.33$ points of average success, so the result does not reduce to having
added a sensor. On a Unitree Z1 with the suction end-effector,
CLAP reaches $96.67\%$ average success over three tasks from one multi-task
checkpoint and places the four-block structure to $15.92$~mm. The cup is still an
actuator; a pressure module costing a few dollars simply lets the arm know what
it holds. Monitoring a cup otherwise needs intricate embedded sensors that are
hard to scale~\cite{hao2026graspguard}. Our contributions can be summarised as
follows:
{\setlength{\topsep}{2pt}%
\begin{itemize}\setlength{\itemsep}{1pt}\setlength{\parsep}{0pt}
  \item \textbf{A contact-sensing path that is complete end to end.} The suction
  state becomes an observable of the system: it enters the policy's
  proprioception in place of the action command usually recorded there, takes
  part in visual reasoning through cross-modal attention, and serves as the abort
  condition of an action block at execution time. The three ablations that touch
  this path differ from the full system by $5.00$ to $11.67$ points.
  \item \textbf{Real-robot goal-state reversal with targeted phase coverage.}
  We record goal-state disassembly on a physical robot and convert the reversed
  absolute joint states directly into next-state targets, with no rollout to
  regenerate. Reversal puts a demonstration's most accurate frame where accuracy
  is needed most; forward, that moment is the beginning, before drift
  accumulates. A small set of phase demonstrations, $8.3\%$ of the training
  frames, covers the suction transitions and the states an interrupted grasp
  leaves behind.
  \item \textbf{Real-robot validation and attribution.} We evaluate across three
  task lengths, two visual settings and both task regimes, attributing the
  outcome across four ablation settings, $1560$ distinct trials in all.
  
\end{itemize}}

\begin{figure*}[t]
  \centering
  \includegraphics[width=\textwidth]{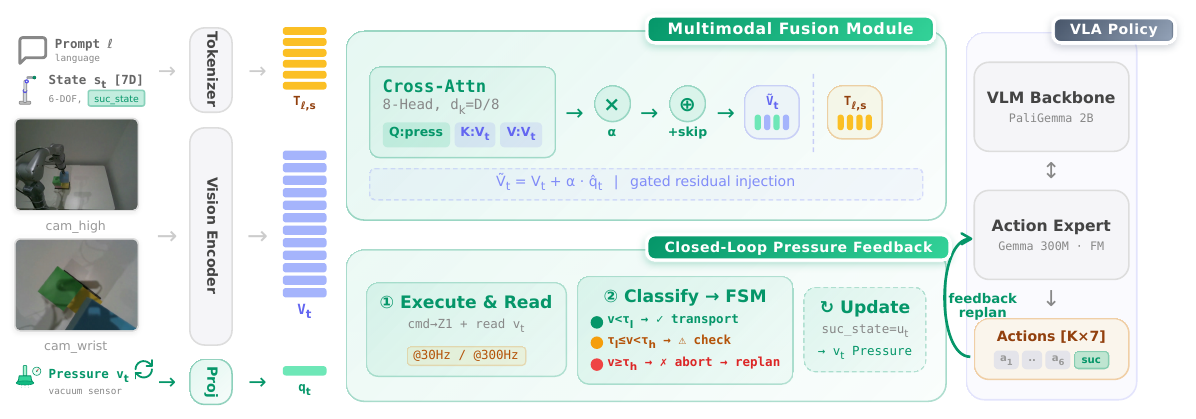}
    \caption{\textbf{Overview of CLAP.} Images and language enter the pretrained
  vision-language backbone. The pressure reading becomes a query attending over
  the visual tokens, the result added back through a gated residual before the
  action expert predicts a block of actions. The same reading replaces the
  suction command in the proprioceptive state, and during execution contradicting
  it aborts the remaining actions and triggers a fresh inference.}
  \label{fig:pipeline}
\end{figure*}

\section{Related Work}
\label{sec:related}

\subsection{Vision-Language-Action Models}
VLA models put perception and control
in one web-pretrained vision-language backbone, so a policy takes language and
generalises to unseen objects
\cite{pmlr-v229-zitkovich23a,kim24openvla,octo_2023,open_x_embodiment_rt_x_2023}.
Their
flow-matching and diffusion heads give smoother trajectories than discrete
tokenisation
\cite{BlackK-RSS-25,intelligence2025pi05visionlanguageactionmodelopenworld,chi2023diffusion}.
The family has also grown in capacity
\cite{team2025gemini,bjorck2025groot}, cross-embodiment reach
\cite{zheng_x-vla_2025} and proprioceptive
input \cite{wang2024hpt,li2024robovlms}. Their reported benchmarks are
dominated by coarse pick-and-place rather than tight alignment, and their loop
has one shape: it closes
through vision and language, and the observation says where the end-effector is,
not whether the contact has happened.

\subsection{Contact Sensing inside VLAs}
Touch is the obvious way to supply what
vision cannot see: tactile skins report contact geometry, force and slip
\cite{yuan2015gelsight,lambeta2020digit}, self-supervised pretraining makes
those representations transferable
\cite{higuera2024sparsh,guzey2023dexterity}, and visuo-tactile embeddings help
where contact must be tracked \cite{huang3dvitac}. Putting touch
inside a VLA is active work. Examples include dynamics-aligned fusion \cite{huang2026tafvla},
dual-path encoding \cite{li2025omnivtla}, contact-gated injection
\cite{zhang2026tacvla}, predictive tactile models \cite{ye2025dreamtacvla},
preference learning \cite{zhang2025vtla}, force-domain diffusion
\cite{wu2024tacdiffusion}. These sensors report the mechanics of a contact that
exists, and are mostly mounted on fingers or wrists, while a suction failure is
decided at the cup--face interface. That failure is the absence of a seal, which
vacuum reports directly, from a component the end-effector already carries. We
report no tactile baseline: the failure modes differ, and adding one changes
end-effector, sensor and task interface at once.

\subsection{Suction in Robot Learning}
Suction is studied mostly as a perception
problem that asks where the cup should go. Models score candidate points
\cite{mahler2018dexnet3}, synthetic benchmarks scale supervision
\cite{li2024simsuction}, dynamics can be folded in \cite{yang2023dynamograsp},
and generative frameworks now cover vacuum grippers \cite{murali2025graspgen}.
That line stops at contact. A second instruments the cup itself through internal
flow for haptic search \cite{lee2023smartsuctioncup,lee2024hapticsearch},
camera-based, dual-zone and coated designs
\cite{sucktac2025,gong2026flexicup,aoyagi2020bellows}. These are hardware
contributions rather than components integrated into a VLA policy. A third uses vacuum outside the
control loop: an offline failure boundary becomes a planning constraint
\cite{avigal2022gompst}, grasp strength a trajectory constraint
\cite{lee2026graspfailure}, cup health a pre-grasp prediction
\cite{hao2026graspguard}, outlet pressure a diagnostic
\cite{baek2021vacuummonitoring}. The nearest reactive system closes a $15$~Hz
loop around a vacuum gripper, but on grasp quality from force-torque feedback,
and for grasping rather than placing \cite{zhang2023reactivevacuum}. Suction has
entered VLAs too, as a binary flag \cite{zhou2025vacuumvla} or an interchangeable
gripper type \cite{zhang2026gvla}. Neither reports a policy conditioned on a
measured vacuum reading, nor one that aborts an action already in progress on
it.

\subsection{Demonstration Quality and Data Strategy}
Precision depends strongly on
contact-phase demonstrations, where teleoperation drift is greatest. Time
reversal exploits motion that is easier from the goal: Wang et al. invert the
pull-out phase of tight-clearance insertion~\cite{wang2026tightclearance}, while
Auto-E2H reverses filtered easy-task rollouts and adds a few hard-task
examples~\cite{qiao2026reverseadvance}. Both draw their reversed trajectories
from the mechanism that struggles forward. Closest to us, RPM records constrained
sub-tasks from goal configurations and unconstrained ones forward, then
re-executes reversed end-effector references through a controller in
simulation~\cite{kim2026rpm}. We instead record physical-robot disassembly and
convert reversed joint states directly into next-state targets, without
regeneration. Targeted phase demonstrations cover suction transitions and
configurations resembling those after an abort. We therefore treat reversal
and per-segment direction choice as established. Our distinction is their
real-robot realisation and integration with pressure-triggered correction.
Elsewhere the same word denotes an RL curriculum over simulator
resets~\cite{tao2024rfcl}. Complementary methods re-collect around atomic
sub-tasks~\cite{yang2025hdspace} or split approach from
insertion~\cite{wang2026tightclearance}. We retain one trunk and checkpoint and
patch only contact-critical phases.

\section{Method}
\label{sec:method}

\subsection{Hardware: Closing the Loop at the System Level}
\label{sec:hw}
\begin{figure*}[t]
  \centering
  \begin{minipage}[b]{0.361\textwidth}
    \centering
    \includegraphics[width=\linewidth]{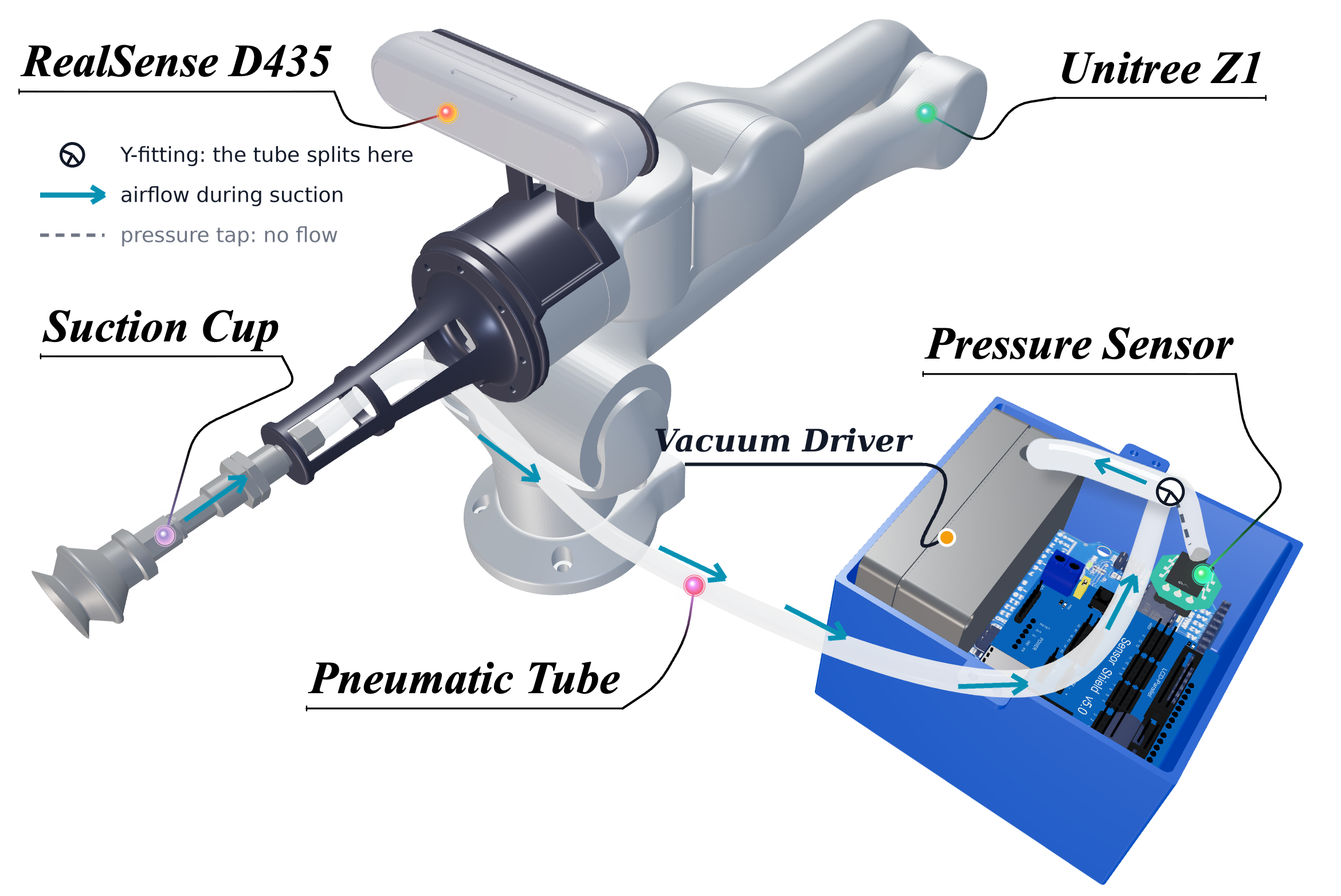}\\[0.25em]
    {\footnotesize (a) end-effector and pneumatic circuit}
  \end{minipage}%
  \hspace{0.04\textwidth}%
  \begin{minipage}[b]{0.470\textwidth}
    \centering
    \includegraphics[width=\linewidth]{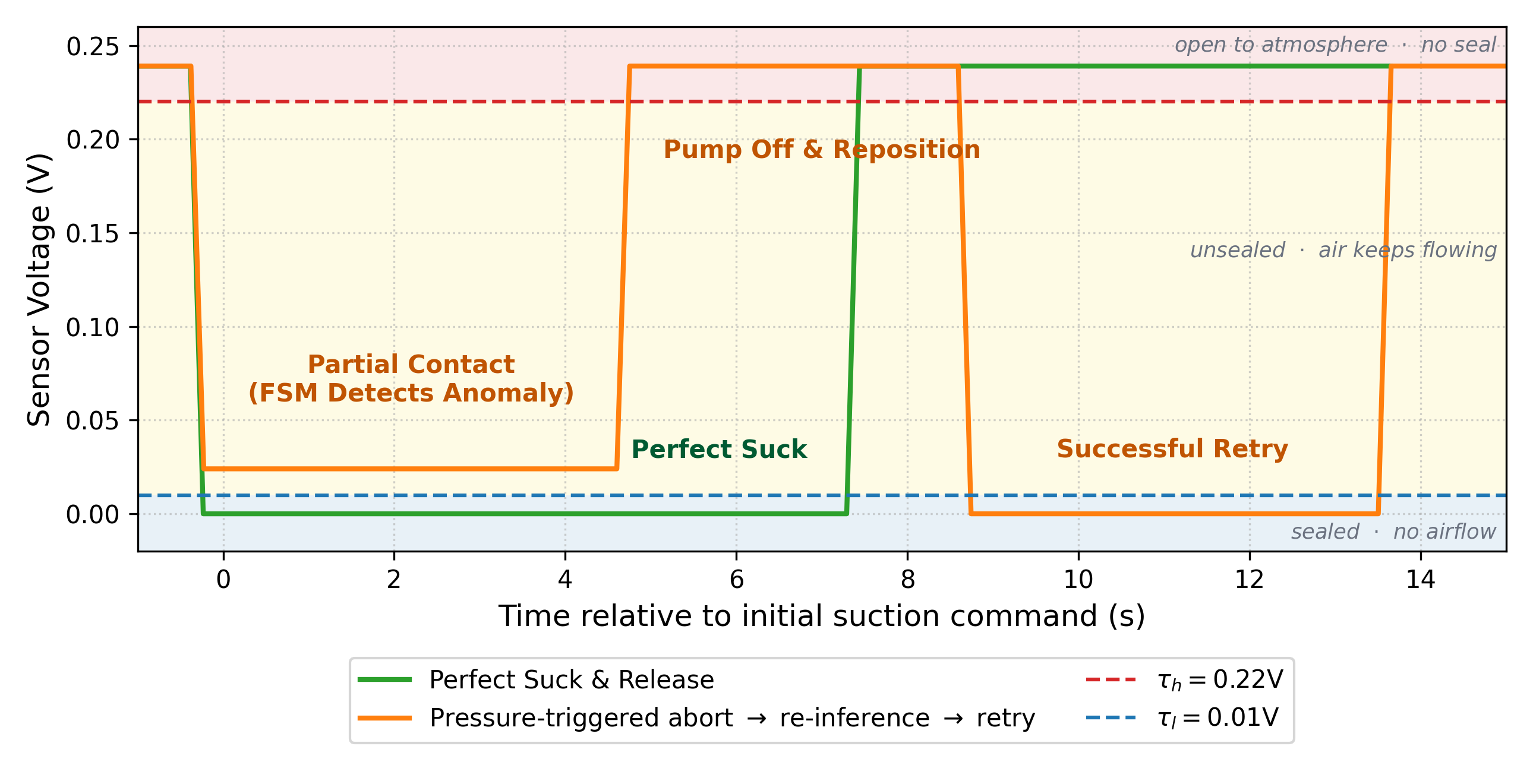}\\[0.25em]
    {\footnotesize (b) vacuum pressure during a sealed and a recovered grasp}
  \end{minipage}
    \caption{\textbf{Hardware and pressure signal.} \textbf{(a)} The end-effector
  and pneumatic circuit. The module sits on the dead-end leg of the
  Y-fitting, so it reads the line's static pressure rather than flow
  (Sec.~\ref{sec:hw}). \textbf{(b)} Decoded contact state over two grasps:
  one seals as the cup meets the face (green), one does not (orange), leaking
  $4.9$~s until a retry seals. The trace is the board's decoded regime, not a
  raw sensor sweep, so $\tau_l$ and $\tau_h$ are shown for reference.}
  \label{fig:hardware}
\end{figure*}

\noindent\textbf{Circuit.} We replace the stock jaw gripper of a Unitree Z1 with a suction end-effector,
and make its state observable by instrumenting the pneumatic circuit rather than
the cup (Fig.~\ref{fig:hardware}a). A tube runs from the cup to a Y-fitting whose
main leg continues to a vacuum driver, which both draws air and breaks the vacuum
on release. The second leg is a dead end holding an XGZP6847A MEMS piezoresistive
module, so the module reports the \emph{static pressure of the line}, not flow
through itself. The module is linear over its range and costs a few dollars, and
nothing is embedded in the cup, which remains an off-the-shelf part.

\noindent\textbf{Three regimes.} Line pressure makes the cup's state well posed because it separates three
\emph{airflow regimes} rather than three arbitrary voltage bands. When sealed
against a face, no air moves and the line reaches the driver's vacuum limit.
When energised but not sealed, air flows through the leak and the line settles
between the two endpoints. Open to the atmosphere, it reads ambient. The middle
regime is one the system occupies rather than passes through. In recorded grasps
it persists $4.9$ and $6.5$~s before a retry
seals, so the thresholds do not bracket a transient. The thresholds $\tau_l = 0.01$~V and $\tau_h = 0.22$~V are
placed just inside the two ends of that range, not at values fitted to a task:
they were not re-tuned across the objects, colours or task lengths here. They do
depend on the circuit. Another pump or cup moves the endpoints and their margins.

An analog voltage leaves the module. The regime it falls in reaches the
policy. An Arduino Uno R3 does that decoding, and the two thresholds
\emph{are} the decoder:
\begin{equation}
\label{eq:contact}
c_t=
\begin{cases}
\mathrm{sealed}  & v_t < \tau_l,\\
\mathrm{leaking} & \tau_l \le v_t < \tau_h,\\
\mathrm{open}    & v_t \ge \tau_h.
\end{cases}
\end{equation}
The latter two both mean that nothing is held, so the policy consumes the binary
$h_t=\mathbf{1}\!\left[c_t=\mathrm{sealed}\right]$. We retain three states because
that is what the signal physically is and because
\emph{leaking} is the regime a retry sits through. Pressure streams over serial
at up to $300$~Hz, synchronised with the D435 images and the Z1 proprioception
at $30$~Hz.

\subsection{Pressure-Aware Policy}
\label{sec:fusion}

\noindent\textbf{The state field.} One field of the proprioceptive state carries the cup,
and it is the only entry of that vector on which the baselines and CLAP
disagree:
\begin{equation}
\label{eq:state}
\mathbf{s}_t=\left[\,\mathbf{q}_t^{\top},\ u_t\,\right]^{\top}\in\mathbb{R}^{7},
\quad
u_t=
\begin{cases}
\text{command} & \text{(baselines)},\\
1-h_t & \text{(ours)},
\end{cases}
\end{equation}
with $\mathbf{q}_t\in\mathbb{R}^{6}$ the joint angles. The baselines read back
their own intent, namely the issued \emph{command} to suck or release. CLAP reads what
the line actually did, in the gripper convention the pretrained model already
uses, where an end-effector holding something reads $0$. The fusion branch below
receives that same $u_t$. The substitution is made at collection time, so the
field means the same thing in training as at deployment, and the state gains no
extra dimension.
\noindent\textbf{Complementarity.} Vision gives the phase, pressure the outcome, and
neither reading is actionable alone: a pick commanded and the cup still
reporting nothing held means the grasp has not taken and the arm should adjust
and retry. Just released and still held means the block is being carried away and
the arm must go back. \emph{Vision knows what is being attempted but not whether
it worked. Pressure knows whether it worked but not what is being attempted.}

\noindent\textbf{Fusion.} Injecting the scalar is not free: concatenated onto hundreds of visual tokens it
is one input among many, a motivation for the branch design that
Sec.~\ref{sec:attr} then prices against the implemented alternative. We lift
$u_t$ into the visual feature space and let it query
the visual tokens. Writing $\mathbf{V}_t\in\mathbb{R}^{L\times D}$ for the tokens the vision
encoder produces,
\begin{align}
\label{eq:proj}
\mathbf{p}_t &= \mathbf{W}_p\,u_t+\mathbf{b}_p \;\in\; \mathbb{R}^{1\times D},\\
\label{eq:attn}
\mathbf{A}^{(h)}_t &= \mathrm{softmax}\!\left(
 \frac{\bigl(\mathbf{p}_t\mathbf{W}^{(h)}_Q\bigr)\bigl(\mathbf{V}_t\mathbf{W}^{(h)}_K\bigr)^{\!\top}}
      {\sqrt{d_k}}\right) \;\in\; \mathbb{R}^{1\times L},\\
\label{eq:mha}
\hat{\mathbf{g}}_t &= \operatorname*{Concat}_{h=1}^{H}
 \bigl(\mathbf{A}^{(h)}_t\,\mathbf{V}_t\mathbf{W}^{(h)}_V\bigr)\,\mathbf{W}_O
 \;\in\; \mathbb{R}^{1\times D},\\
\label{eq:gate}
\tilde{\mathbf{V}}_t &= \mathbf{V}_t+\alpha\,\hat{\mathbf{g}}_t,
 \qquad \alpha_0=0,
\end{align}
with $H=8$ heads and $d_k=D/H$. The shape of $\mathbf{A}^{(h)}_t$ is the whole
argument: a distribution over the $L$ patches, so the contact reading chooses
\emph{which part of the scene} to summarise (Fig.~\ref{fig:pipeline}). The
summary $\hat{\mathbf{g}}_t$ is then a \emph{single} vector added to every row of
$\mathbf{V}_t$. Attention selects the patches, while injection is uniform.
Because $\alpha$ starts at zero, fine-tuning begins from the pretrained
$\pi_{0.5}$\cite{intelligence2025pi05visionlanguageactionmodelopenworld}
computation with the state field carrying the sensor reading. The action
is
\begin{equation}
\label{eq:action}
\mathbf{a}_t=\bigl(\mathbf{a}^{\text{arm}}_t\in\mathbb{R}^{6},\ \tilde{a}^{\text{suc}}_t\bigr),
\quad
a^{\text{suc}}_t=1+\mathbf{1}\!\left[\tilde{a}^{\text{suc}}_t\ge\tfrac{1}{2}\right],
\end{equation}
the cup command relaxed to a scalar that the same flow-matching head regresses
and execution thresholds back onto the two codes $\{1,2\}$ the driver accepts,
a measurement in the state and a command here. Sharing the head is simple, but
it buries a discrete decision, \emph{when} to switch the cup, inside a
continuous regression (Sec.~\ref{sec:data}).

\subsection{Reactive Correction: Breaking the Open-Loop Chunk}
\label{sec:loop}
$\pi_{0.5}$\cite{intelligence2025pi05visionlanguageactionmodelopenworld}
predicts $K=50$ actions and executes the first $25$, which at $30$~Hz is
$0.83$~s of open-loop motion. Whatever happens to the object
inside that window, an ordinary policy finishes the chunk: if the cup lets go
early, the arm flies the rest of the trajectory empty and places nothing.
Chunking is standard across modern
VLAs~\cite{ACT-RSS-23,chi2023diffusion,BlackK-RSS-25} and this exposure a known
property of it. What has been missing is a signal reliable enough to interrupt
on. Vision is badly placed to provide one: the cup and the face it holds occlude
each other exactly when the answer matters. Vacuum is well placed: the seal
either exists or it does not.

\noindent\textbf{Abort condition.} CLAP therefore gives the chunk one: when the cup reports
nothing held under a suction command, execution stops at the current
step, a fresh observation is taken, and a new chunk is inferred. Let $d_k$ count
the consecutive such steps up to $k$. Execution runs to
\begin{equation}
\label{eq:abort}
k^{\star}=\min\bigl\{\,k\le K/2 \;:\; d_k = N\,\bigr\},
\qquad N=4,
\end{equation}
and re-infers from the observation at $t+k^{\star}$. If no such $k$ exists, the
half-chunk is executed in full and correction waits for the next one. One
window covers two failures: a grasp that never takes and a seal lost in transit.
We set $N$ from the physics rather than tuning it. A seal that will form does so within one or two steps, so four
steps, $0.13$~s at $30$~Hz, clears that margin without waiting out the
$0.83$~s chunk. The condition supplies only the \emph{timing}, which determines
when to stop. The policy produces where the arm goes afterwards from the new
observation, so this motion is learned rather than scripted.
Two things outside the loop make that possible. The contact reading is the
genuine one, so \emph{not held} is an ordinary in-distribution observation. It
holds throughout the approach phase of every demonstration, and the
configuration an abort leaves behind is covered by the supplementary segments of
Sec.~\ref{sec:data}. A correction can thus arrive immediately or only once the
chunk has run out, and Sec.~\ref{sec:attr} measures the difference. At
millimetre scale, noticing failure is worth little once the arm has moved on.

Force and torque sensing could also report a lost grasp because a wrist sensor
registers the load added by a held block. Contact-rich assembly commonly relies
on this signal~\cite{wang2026tightclearance,wu2024tacdiffusion}. Vacuum instead
reads the seal itself from a component the end-effector already
carries.
\subsection{Data Strategies}
\label{sec:data}

\begin{figure}[t]
  \centering
  \includegraphics[width=0.92\columnwidth]{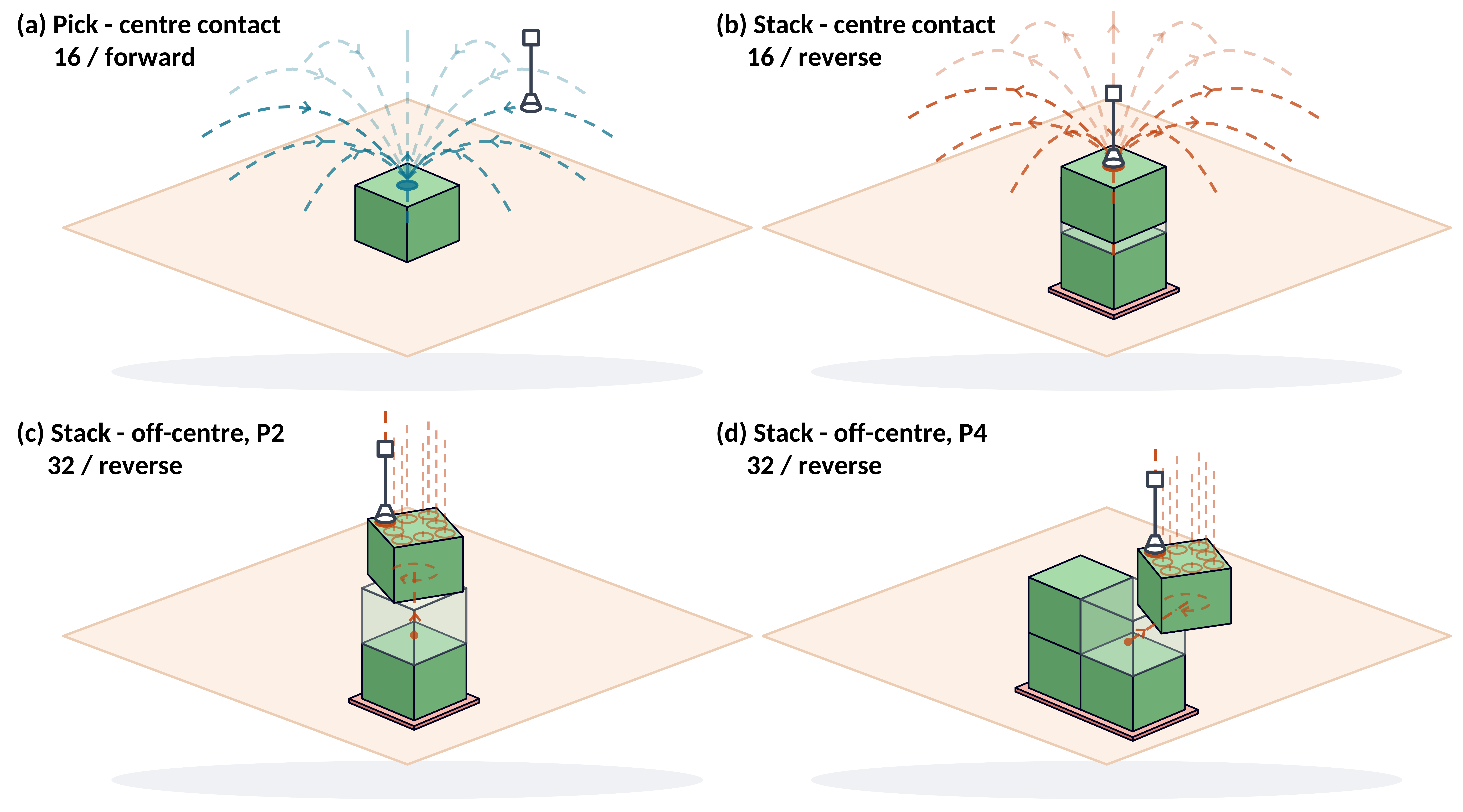}
    \caption{\textbf{The supplementary segments.} Drawn to scale. \emph{Top}, the
  cup contacts the centre of the top face: \textbf{(a)} a pick and \textbf{(b)} a
  stacking placement, each recorded from many directions and starting positions.
  \emph{Bottom}, the same placement with the contact point moved off centre until
  the rim reaches the edge, for \textbf{(c)} $P_2$ and \textbf{(d)} $P_4$. Arrows
  give the recording direction.}
  \label{fig:data}
\end{figure}

\begin{figure}[t]
  \centering
  \includegraphics[width=\columnwidth]{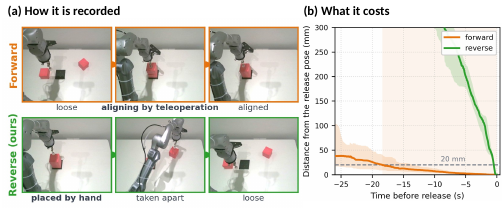}
    \caption{\textbf{Reverse-order acquisition.} \textbf{(a)} The same $P_2$ task
  recorded both ways from the head camera. The two bold labels mark the one
  thing the frames cannot show, which direction produced the alignment.
  \textbf{(b)} Distance from the release pose against time before release. Median
  and inter-quartile range, with the forward dwell shaded.}
  \label{fig:reverse}
\end{figure}

Both data strategies follow from one observation: the precision the policy must
learn is concentrated in the last few millimetres of each placement, and that is
exactly where a teleoperated demonstration is worst.

\noindent\textbf{Reverse-order acquisition.} Rather than demonstrate the assembly, the
operator builds the finished structure \emph{by hand} before recording starts,
setting the terminal pose directly rather than steering to it, and records only
the disassembly. Reversing it in time
yields a training episode (Fig.~\ref{fig:reverse}\textcolor{red}{a}). The hand contributes the
accuracy of the final pose, rather than the rig. No part of the episode was
produced by aligning the object through the leader arm. The disassembly itself
is still teleoperated. Time reversal has been used for the fine-insertion phase of a
hierarchical pipeline~\cite{wang2026tightclearance} and, closest to us,
for constrained sub-tasks recorded from the goal
configuration~\cite{kim2026rpm}. In our multi-block setting a long horizon adds
two constraints. The disassembly must strictly invert the stacking order, and
the occlusion relations run backwards. We therefore plan the demonstration as a whole.
Reversal is cheap. If the disassembly is recorded as
$\mathbf{q}_0\ldots\mathbf{q}_T$, the offline conversion writes, for the arm,
\begin{equation}
\label{eq:reverse}
\tilde{\mathbf{q}}_t=\mathbf{q}_{T-t},
\qquad
\mathbf{a}^{\text{arm}}_t=\tilde{\mathbf{q}}_{t+1},
\end{equation}
an absolute next-joint target read off the reversed sequence rather than a
negated increment. The usual training transform then turns these six dimensions
into displacements relative to the current state. Recomputing them from the
reversed sequence gives the correct sign without manual flipping. Nothing is
flipped by hand. The suction channel is left absolute: a grasp and a release
being each other's time reverse, the recorded codes can be reused as they are.

The recordings show what this buys (Fig.~\ref{fig:reverse}\textcolor{red}{b}). Collecting $P_2$
forward took one operator two days. The same number of reverse episodes took half
a day. Per placement the gap is sharper. Write $\tau_{\mathrm{dwell}}$ for the
time the end-effector spends inside a ball of radius $20$~mm around the point
where it finally lets go, counted back from release at $30$~Hz. Forward
demonstrations spend a median $\tau_{\mathrm{dwell}}$ of $14.6$~s for the first
block and $26.4$~s for the second, hunting for the alignment, against $0.5$~s in
reverse. These are medians over
$100$ forward and $100$ reverse episodes. Those intervals are about $40\%$ of a
forward episode (median over episodes) spent hesitating at a target, and nothing
in a behaviour-cloning objective penalises that. The second block is worse than
the first because it aligns to the first rather than to the mat. Forward collection of
$P_4$ was never attempted. What we measure here is the demonstrations, not the
policies they produce: every model in this paper trains on the same corpus.

\noindent\textbf{Direction is chosen per segment.} A full trajectory is always reversed
because terminal alignment is the bottleneck. A short segment is recorded in
whichever direction reaches the required accuracy. Picking a block off the table
is recorded forward, while every stacking segment is reversed. RPM exploits the
same asymmetry between constrained and unconstrained segments~\cite{kim2026rpm}. Of the $64$ supplementary segments
per task, $16$ are picks and $48$ placements. The picks begin with the cup empty
and a block on the table, like the configurations a missed grasp leaves behind,
so recovery
after an abort is a situation the policy has been trained in rather than one it
must invent~\cite{yang2025hdspace}.

\noindent\textbf{Minimal phase patches.} The segments are short, about a quarter of a full
episode, and few: $192$ trajectories, $8.3\%$ of all training frames. They vary
along two crossed axes (Fig.~\ref{fig:data}). The first is the step in isolation,
a pick or a placement alone, which lets the flow-matching head separate the
discrete pneumatic decision from the continuous arm motion it is regressed
alongside. This difficulty comes from the shared head of Sec.~\ref{sec:fusion}
rather than the task, although we do not test that motivation in isolation. The second is the contact point: half the segments centre the cup on the top
face, half move it off centre until the rim reaches the edge, forcing the policy
to compensate for the resulting tilt in transport. Approach direction and layout
vary across every segment. The patches supplement the full trajectories rather
than replacing them, so one checkpoint still covers the task.

\section{Experiments}
\label{sec:exp}

\subsection{Hardware and Experimental Setup}
\label{sec:setup}

\begin{wrapfigure}{r}{0.5\columnwidth}
  \vspace{-\intextsep}
  \centering
  \includegraphics[width=0.48\columnwidth]{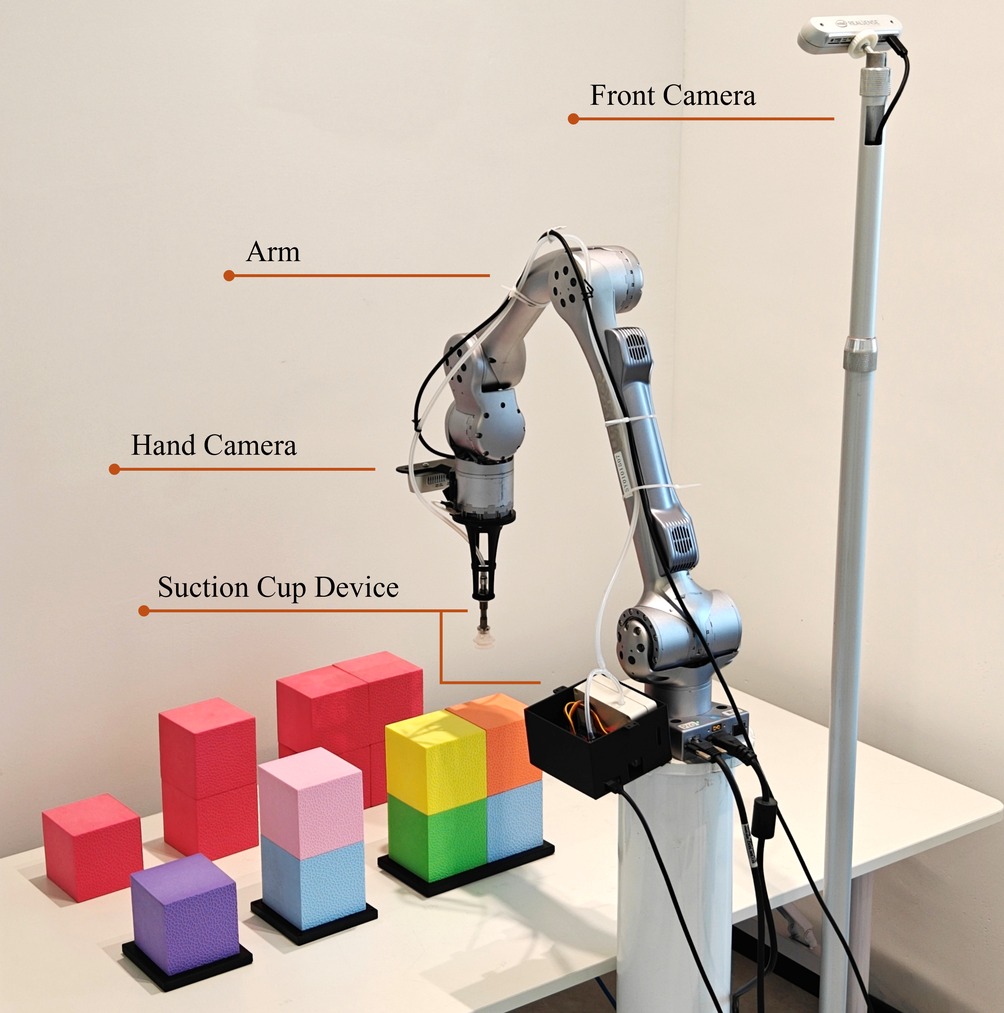}
  \caption{\textbf{Real-world task setting.}}
  \label{fig:setup}
  \vspace{-\intextsep}
\end{wrapfigure}

\noindent\textbf{Platform.} All experiments run on a Unitree Z1 with the suction end-effector of
Sec.~\ref{sec:hw} and two RealSense D435 cameras, third-person and wrist
(Fig.~\ref{fig:setup}).

\noindent\textbf{Tasks.} $P_1$ places one block on a pad, $P_2$ stacks two, $P_4$ builds two such stacks
side by side: a volumetric $2\times2$ of four blocks. Blocks are
$80\times80\times80$~mm, pads $100\times100$~mm for $P_1$, $P_2$ and
$180\times100$~mm for $P_4$, so the first layer clears the pad edge by
$\pm10$~mm, an eighth of a block edge, in every task and direction. The longer
tasks are not held to a tighter tolerance but must meet it repeatedly, in $P_4$
along two axes at once.

\noindent\textbf{Randomisation.} Blocks and pad are randomised in position and yaw over the whole reachable
tabletop, for collection and evaluation alike. Yaw spans the full circle, but a policy
must null only the residual modulo the object's symmetry: a cubic block and the
square pads of $P_1$, $P_2$ are four-fold symmetric, so it never exceeds
$45^\circ$ and every alignment is correct. $P_4$ differs in kind: its
$180\times100$~mm pad is two-fold symmetric, the residual runs to $90^\circ$
and the long axis must be identified: two blocks side by side occupy $160$~mm,
clearing the long side by $\pm10$~mm and overflowing the short: reading
the pad $90^\circ$ wrong does not cost accuracy, it makes the layout
infeasible.

Forward kinematics, agreeing with the arm's own pose reports to $0.00$~mm once
its constant tool offset is applied, gives the end-effector position at every
contact event in all $567$ episodes. Over those $1942$ events the engage and release poses cover
$360\times755$~mm of table, $2374$~cm$^2$, and a block travels from $97$ to
$683$~mm between pick-up and set-down, median $270$~mm. Nothing shrinks as the task lengthens: $P_1$ spans
$338\times726$~mm, $P_2$ $337\times712$, $P_4$ the largest at $360\times755$.
Collection sweeps that space coarsely. The $20$ evaluation poses per setting sit
\emph{between} the training poses, including near-edge cases and configurations
where blocks touch each other or the pad. The pad's pose must be resolved too, so
the objects to localise number one more than the blocks, with three for $P_2$
and five for $P_4$.

\noindent\textbf{Protocol.} A trial succeeds if no first-layer block overhangs its pad, the structure stands
$5$~s, and the largest distance between corner pairs that should coincide stays
within $20$~mm for $P_2$ and $40$~mm for $P_4$. These bounds equal a quarter and
a half of a block edge, the looser for the task with more corner pairs. We read
that distance off vernier callipers, a completed stack occluding the corners an
image-based estimate would need. We report
success over $20$ trials per setting and, for successes, that maximum as
mean\,$\pm$std. Across both tables that is $1560$ distinct trials. Each task runs with blocks of one colour and of several. The
instruction is identical and never names colour, and both use one checkpoint
trained on both appearances, so colour changes only how readily a crowded scene
separates into blocks.

\noindent\textbf{Data and training.} The corpus is $567$ trajectories, $196$~min recorded at $30$~Hz and
temporally subsampled to $146{,}774$ training frames: $375$ full-task
trajectories collected in reverse order and $192$ segments of
Sec.~\ref{sec:data}. Training starts from the released $\pi_{0.5}$ checkpoint and
updates all parameters at batch size $96$ over four devices, EMA disabled, on a
cosine schedule with $1$k warmup peaking at $2.5\mathrm{e}{-5}$ and decaying to
$2.5\mathrm{e}{-6}$ over the $30$k steps at which the evaluated checkpoints are
taken. The single-task model sees one task's episodes, the multi-task all three.
Every baseline trains on the same data, sees the same images, and is scored by
the same criterion.

\subsection{Comparison with State-of-the-Art}
\label{sec:cmp}


\begin{table*}[t]
\centering
\providecommand{\pmstd}[1]{\textcolor{gray}{\,$\pm$#1}}
\caption{\textbf{Hierarchical stacking tasks.} SR in percent; alignment error in
mm as mean\,\textcolor{gray}{$\pm$std} over successful trials, the maximum
distance between corner pairs that should coincide. N/A: alignment error is not
reported below a $25\%$ success rate.}
\label{tab:main}
\setlength{\tabcolsep}{4pt}
\resizebox{\textwidth}{!}{%
\begin{tabular}{l l cccccc cccccc}
\toprule
 & & \multicolumn{6}{c}{\textbf{Monochromatic}} & \multicolumn{6}{c}{\textbf{Polychromatic}} \\
\cmidrule(lr){3-8}\cmidrule(lr){9-14}
\multirow{2}{*}{\textbf{Setting}} & \multirow{2}{*}{\textbf{Method}}
 & $P_1$ & \multicolumn{2}{c}{$P_2$} & \multicolumn{2}{c}{$P_4$} & \multirow{2}{*}{Avg}
 & $P_1$ & \multicolumn{2}{c}{$P_2$} & \multicolumn{2}{c}{$P_4$} & \multirow{2}{*}{Avg} \\
\cmidrule(lr){3-3}\cmidrule(lr){4-5}\cmidrule(lr){6-7}
\cmidrule(lr){9-9}\cmidrule(lr){10-11}\cmidrule(lr){12-13}
 & & SR & SR & Err & SR & Err & & SR & SR & Err & SR & Err & \\
\midrule
\multirow{7}{*}{\textit{Single-task}}
 & ACT~\cite{ACT-RSS-23}          & 30 & 0 & N/A & 0 & N/A & 10.00 & 15 & 0 & N/A & 0 & N/A & 5.00 \\
 & DP~\cite{chi2023diffusion}           & 45 & 0 & N/A & 0 & N/A & 15.00 & 20 & 0 & N/A & 0 & N/A & 6.67 \\
 & DP3~\cite{ze20243d}          & 20 & 0 & N/A & 0 & N/A & 6.67  & 20 & 0 & N/A & 0 & N/A & 6.67 \\
 & X-VLA~\cite{zheng_x-vla_2025}        & 60 & 65 & 13.29\pmstd{2.82} & 45 & 23.82\pmstd{5.38} & 56.67 & 55 & 60 & 13.94\pmstd{2.05} & 45 & 21.77\pmstd{2.87} & 53.33 \\
 & $\pi_0$~\cite{BlackK-RSS-25}      & 35 & 30 & 14.87\pmstd{4.16} & 10 & N/A & 25.00 & 30 & 30 & 12.92\pmstd{4.12} & 5 & N/A & 21.67 \\
 & $\pi_{0.5}$~\cite{intelligence2025pi05visionlanguageactionmodelopenworld}  & 75 & 60 & 12.53\pmstd{4.32} & 50 & 23.06\pmstd{10.12} & 61.67 & 65 & 60 & 13.64\pmstd{4.20} & 55 & 20.59\pmstd{2.50} & 60.00 \\
 & \textbf{Ours}& \textbf{100} & \textbf{90} & \textbf{11.12}\pmstd{4.31} & \textbf{80} & \textbf{14.62}\pmstd{3.35} & \textbf{90.00}
                & \textbf{95}  & \textbf{80} & \textbf{9.54}\pmstd{3.29}  & \textbf{70} & \textbf{17.31}\pmstd{7.47} & \textbf{81.67} \\
\midrule
\multirow{4}{*}{\shortstack[l]{\textit{Multi-task}\\[-1pt]\scriptsize(one checkpoint)}}
 & X-VLA~\cite{zheng_x-vla_2025}        & 80 & 85 & 10.72\pmstd{2.87} & 75 & 17.78\pmstd{3.76} & 80.00 & 95 & 85 & 10.27\pmstd{3.20} & 80 & 19.06\pmstd{2.78} & 86.67 \\
 & $\pi_0$~\cite{BlackK-RSS-25}      & 50 & 45 & 13.32\pmstd{4.02} & 20 & N/A & 38.33 & 85 & 60 & 10.55\pmstd{3.70} & 30 & 22.87\pmstd{3.02} & 58.33 \\
 & $\pi_{0.5}$~\cite{intelligence2025pi05visionlanguageactionmodelopenworld}  & 75 & 75 & 10.47\pmstd{4.23} & 60 & 18.74\pmstd{5.25} & 70.00 & 90 & 80 & 9.38\pmstd{3.77}  & 70 & 18.86\pmstd{3.86} & 80.00 \\
 & \textbf{Ours}& \textbf{100} & \textbf{100} & \textbf{8.42}\pmstd{3.92} & \textbf{90} & \textbf{15.92}\pmstd{4.32} & \textbf{96.67}
                & \textbf{100} & \textbf{100} & \textbf{7.50}\pmstd{2.82} & \textbf{90} & \textbf{13.51}\pmstd{3.68} & \textbf{96.67} \\
\bottomrule
\end{tabular}}
\end{table*}

Table~\ref{tab:main} separates two regimes. The task-specific models fall off a
cliff as soon as more than one block is involved: ACT~\cite{ACT-RSS-23},
Diffusion Policy~\cite{chi2023diffusion} and DP3~\cite{ze20243d} place a single
block with some success but reach $0\%$ on both $P_2$ and $P_4$,
in either colour setting. With every object anywhere in that $2374$~cm$^2$
footprint at an arbitrary yaw, and the pad moving too, the policy must first
decide \emph{which} object to approach -- a problem $P_1$ hides and the VLA
baselines do not have.
$\pi_{0.5}$\cite{intelligence2025pi05visionlanguageactionmodelopenworld} and
X-VLA~\cite{zheng_x-vla_2025} carry pretrained spatial priors into the
multi-block tasks and get the arm to the right place, then stall at the last
mile. Multi-task monochromatic, the strongest baseline X-VLA reaches $80.00\%$
and $\pi_{0.5}$ $70.00\%$, the latter at $10.47$ and $18.74$~mm on $P_2$ and
$P_4$, against CLAP's $96.67\%$ at $8.42$ and $15.92$~mm. The gap is in
finishing, not in reaching.

CLAP's alignment error is \emph{lower} in the multi-task setting than in the
single-task one, $8.42$ against $11.12$~mm on $P_2$, although one checkpoint
covers all three tasks: the supplementary segments are shared, so each task sees
more contact-phase supervision than it contributed. The improvement is in spread
as much as in mean, the standard deviation on single-task $P_4$ being $3.35$~mm
for CLAP against $10.12$~mm for $\pi_{0.5}$, and spread is what makes a stack
collapse when the next block goes on top. The polychromatic column moves the
baselines up, X-VLA to $86.67\%$, while CLAP stays at $96.67\%$; multi-task
training helps every method, CLAP least at $+6.67$ points.

\subsection{Attribution and Ablation}
\label{sec:attr}
\label{sec:abl}

\begin{table}[t]
\centering
\caption{\textbf{Component ablation.} Multi-task monochromatic setting, same
protocol as Table~\ref{tab:main}. Variants are defined in Sec.~\ref{sec:abl}.
$\Delta$ is the drop from the full model.}
\label{tab:ablation}
\setlength{\tabcolsep}{3.5pt}
\resizebox{\linewidth}{!}{%
\begin{tabular}{l ccc cc cc}
\toprule
\multirow{2}{*}{\textbf{Variant}} & \multicolumn{3}{c}{SR (\%)} &
\multirow{2}{*}{\textbf{Avg}} & \multirow{2}{*}{$\Delta$} &
\multicolumn{2}{c}{Err (mm)} \\
\cmidrule(lr){2-4}\cmidrule(lr){7-8}
 & $P_1$ & $P_2$ & $P_4$ & & & $P_2$ & $P_4$ \\
\midrule
\rowcolor{gray!12}
\textbf{Ours (full)}        & \textbf{100} & \textbf{100} & \textbf{90} & \textbf{96.67} & 0.00    & \textbf{8.42}  & \textbf{15.92} \\
\midrule
w/o real-time interruption  & 100 & 90 & 85 & 91.67 & $-5.00$  & 10.79 & 18.48 \\
w/o modal fusion            & 90  & 90 & 80 & 86.67 & $-10.00$ & 10.72 & 17.78 \\
w/o pressure feedback       & 95  & 85 & 75 & 85.00 & $-11.67$ & 10.68 & 18.04 \\
w/o sub-task patches        & 90  & 85 & 75 & 83.33 & $-13.33$ & 13.39 & 20.35 \\
w/o all ($\pi_{0.5}$)       & 75  & 75 & 60 & 70.00 & $-26.67$ & 10.47 & 18.74 \\
\bottomrule
\end{tabular}}
\end{table}

\noindent\textbf{Variants.} All variants share the pretrained weights, the reverse-order backbone and the
schedule. \emph{w/o real-time interruption} disables only immediate abort, so
a correction waits for the chunk to end;
\emph{w/o pressure feedback} restores command semantics, zeros the branch and
also lacks abort; \emph{w/o modal fusion} keeps feedback and abort but
concatenates the scalar instead of attending over it; \emph{w/o sub-task
patches} removes the $192$ segments; \emph{w/o all} is multi-task
$\pi_{0.5}$.

\noindent\textbf{Two controlled steps.} Disabling immediate abort lowers average success from $96.67\%$ to $91.67\%$.
These variants differ only in when correction may occur, so the $5.00$-point
gap isolates reacting within a chunk rather than after it. Comparing \emph{w/o
real-time interruption} with \emph{w/o pressure feedback}, with abort absent in
both, isolates observed feedback at fixed data, parameter count and schedule, at
a further $6.67$ points. Both steps are three and four successes in sixty, an
ordering rather than a measurement. Over the pooled $60$ trials a two-sided
Fisher exact test resolves the drops to \emph{w/o sub-task patches} ($p=0.03$)
and to $\pi_{0.5}$ ($p<0.001$); the $10.00$ points against concatenation fall in
between, at $p=0.09$. The abort supplies
only timing.
Recovery motion comes from the newly inferred policy chunk. We do not decompose
the remaining distance to $\pi_{0.5}$, which also lacks the fusion design and
phase segments. Variants are retrained, not switched at inference.

\noindent\textbf{Fusion and phase coverage.} With feedback and abort retained, the scalar in \emph{w/o modal fusion} is present, synchronised and
correct throughout training, yet concatenation reaches $86.67\%$ versus
$96.67\%$ with cross-attention. Its $1.67$-point gap over \emph{w/o pressure
feedback}, and that setting's $11.67$-point drop, are both composite, because
\emph{w/o pressure feedback} also lacks abort. Removing only the $192$ phase
segments, $8.3\%$ of all frames, gives $83.33\%$ and the largest alignment
errors ($13.39$ and $20.35$~mm), consistent with their concentration around
contact. The four drops are not additive ($40.00$ against $26.67$ for \emph{w/o
all}) and range from $5.00$ to $13.33$ points, the clean largest belonging to the
phase segments rather than to the sensor. The lowest success rate is not the
worst error.

All of this is interpolation across the sampled workspace
(Sec.~\ref{sec:setup}). We omit a parallel-jaw baseline: changing the end-effector changes the embodiment,
reachable object set and failure modes together, isolating nothing.

\subsection{Failure Modes and Boundaries}
\label{sec:fail}

\noindent\textbf{Failure categories.} We score each failed trial as a block over a boundary or noticeably rotated (C),
stacked in the wrong order (W), a placed block knocked out of position (E), a
trajectory finished empty-handed after a missed grasp (N), or wandering without
localising a target (F). (W) presupposes an order, and so two blocks. A failed
attempt followed by a successful one counts as a success. The two baseline families fail differently. For the task-specific
models the characteristic failure is (F): the scene they must resolve is never
the scene they memorised.
The VLA baselines localise, and fail instead as (N): with no contact signal a
missed grasp goes unnoticed and the rest of the trajectory is carried out on
nothing.

\noindent\textbf{CLAP's own boundaries.} A correction can be destructive when the
arm repositions into a block already placed (E), and an off-centre grasp not
compensated during transport shows up as (C). In $P_4$, under heavy occlusion
from the completed stacks, the backbone sometimes treats the task as finished and
stops early. This failure of temporal context is invisible to pressure feedback
because nothing is wrong with the contact. The loop sees contact, not progress.

\section{Conclusion}
\label{sec:conclusion}
\textbf{CLAP} closes the loop on suction attachment during precise placement. A pressure
module tapped into the vacuum line makes the attachment state observable: the
decoded reading replaces the commanded suction state in proprioception, is fused
with the visual features, and is the condition on which an open-loop action
block is abandoned for a fresh inference. The data is goal-state disassembly
recorded on the physical robot and reversed offline without a simulation replay,
plus targeted phase demonstrations that comprise $8.3\%$ of the frames and cover the suction transitions and
the configurations an interrupted grasp leaves behind. Across three tasks one
multi-task checkpoint averages $96.67\%$ success in both colour
settings, and the four ablation settings lie $5.00$ to $13.33$ points below it,
none of them back at the $70.00\%$ baseline. Nothing sits inside the cup, so the
same line tap suits another vacuum end-effector once its circuit is
re-characterised. Isolating the policy-level effect of reversal needs a
forward-against-reverse comparison at matched cost.

\bibliographystyle{IEEEtran}
\bibliography{egbib}

\end{document}